\documentclass[conference]{IEEEtrans}
\IEEEoverridecommandlockouts
\usepackage[margin=0.75in]{geometry}
\usepackage{cite}
\usepackage{amsmath,amssymb,amsfonts}
\usepackage{algorithmic}
\usepackage{graphicx}
\usepackage{textcomp}
\usepackage{xcolor}
\usepackage{tabularx}
\usepackage{authblk}
\usepackage{booktabs}
\usepackage{multirow}
\usepackage{siunitx}
\usepackage{caption}
\usepackage{paralist}
\usepackage{svg}
\usepackage{subcaption}
\usepackage{scrextend}
\usepackage[colorlinks=true]{hyperref}

\def\BibTeX{{\rm B\kern-.05em{\sc i\kern-.025em b}\kern-.08em
    T\kern-.1667em\lower.7ex\hbox{E}\kern-.125emX}}
\begin{document}
\bibliographystyle{IEEEtran}

\newcommand{\todo}[1]{\textbf{\textcolor{red}{TODO: #1}}}

\renewcommand{\IEEEtitletopspaceextra}{20pt}

\title{CLIPGraphs: Multimodal Graph Networks 
to Infer Object-Room Affinities 
% VCSP: Visual Common-Sense Priors\\
\thanks{*Denotes equal contribution}
}

\author{Ayush Agrawal$^{*1}$, Raghav Arora$^{*1}$, Ahana Datta$^{1}$, Snehasis Banerjee$^{1,2}$, Brojeshwar Bhowmick$^2$, \\ Krishna Murthy Jatavallabhula$^{3}$, Mohan Sridharan$^4$, Madhava Krishna}
\affil[1]{Robotics Research Center, IIIT Hyderabad, India}
\affil[2]{TCS Research, Tata Consultancy Services, India}
\affil[3]{CSAIL, Massachusetts Institute of Technology, USA}
\affil[4]{Intelligent Robotics Lab, University of Birmingham, UK}

\renewcommand\Authands{ and }

\maketitle

%%%%%%%%%%%%%%%%%%%%%%%%%%%%%%%%%%%%%%%%%%%%%%%%%%%%%%%%%%%%%%%%%%%%%
%%%%%%%%%%%%%%%%%%%%%%%%%%%%%%%%%%%%%%%%%%%%%%%%%%%%%%%%%%%%%%%%%%%%%
\begin{abstract}
This paper introduces a novel method for determining the best room to place an object in, for embodied scene rearrangement.
% Computing the most suitable room for any given object is an important step in the scene rearrangement challenge for embodied AI.
While state-of-the-art approaches rely on large language models (LLMs) or reinforcement learned (RL) policies for this task, our approach, CLIPGraphs, efficiently combines commonsense domain knowledge, data-driven methods, and recent advances in multimodal learning.
Specifically, it (a) encodes a knowledge graph of prior human preferences about the room location of different objects in home environments, (b) incorporates vision-language features to support multimodal queries based on images or text, and (c) uses a graph network to learn object-room affinities based on embeddings of the prior knowledge and the vision-language features.
We demonstrate that our approach provides better estimates of the most appropriate location of objects from a benchmark set of object categories in comparison with state-of-the-art baselines\footnote{Supplementary material and code: \href{https://clipgraphs.github.io}{https://clipgraphs.github.io}}.
% We propose a multimodal learning approach to infer human preferences of a \emph{tidy room}. This is crucial for robotic scene rearrangement problems, where much of the prior work has centered around assuming rearrangement goals to be concretely specified.
% Our method, dubbed CLIPGraphs, encodes features from off-the-shelf vision-language models (specifically CLIP) using a graph neural network over the set of rooms and objects present in a scene.
% CLIPGraphs are multimodal -- they may be queries using both vision and language queries, and map each query to a room of the house the object should belong to, in line with human preferences.

\end{abstract}

\begin{IEEEkeywords}
Commonsense knowledge, graph convolutional network, knowledge graph, large language models, scene rearrangement.
\end{IEEEkeywords}
%%%%%%%%%%%%%%%%%%%%%%%%%%%%%%%%%%%%%%%%%%%%%%%%%%%%%%%%%%%%%%%%%%%%%
%%%%%%%%%%%%%%%%%%%%%%%%%%%%%%%%%%%%%%%%%%%%%%%%%%%%%%%%%%%%%%%%%%%%%

%%%%%%%%%%%%%%%%%%%%%%%%%%%%%%%%%%%%%%%%%%%%%%%%%%%%%%%%%%%%%%%%%%%%%
%%%%%%%%%%%%%%%%%%%%%%%%%%%%%%%%%%%%%%%%%%%%%%%%%%%%%%%%%%%%%%%%%%%%%
\section{Introduction}
 % Robotic agents are becoming increasingly prevalent in our daily lives, performing tasks that are too dangerous, difficult, or time-consuming for humans. \textbf{Can skip this}.
 %

 % This allows the agent to perform a wide range of tasks from fetching multiple objects in a house to assisting in medical procedures.
 %

% There have been recent developments in Embodied AI on the use of commonsense reasoning for an agent to perform tasks without explicit instructions. 

Imagine a robot being tasked with tidying up an unfamiliar house. This task is a variant of the \textit{scene rearrangement} challenge for embodied AI~\cite{batra2020rearrangement}.
To perform this task, the robot must first determine what tidying up means in this specific house, which requires constructing a representation of the current state of the house and inferring a possible goal state (i.e., a configuration in which the house is deemed \emph{tidy}).
% To perform this task, the robot has to infer what \emph{tidying up} implies in this house. This, in turn, entails the construction of a representation of the current state of the house and the goal state that corresponds to the house after it has been tidied up.
Any errors in this step can influence downstream planning and control, resulting in irrecoverable failure.
% compute and execute actions that change the current state to the goal state.
Computing the most appropriate room location for specific object categories is thus critical to the successful completion of such tasks. %While much of the existing work assumes that a goal state is already determined by a human operator~\cite{goodwin2021semantically, Weihs2021VisualRR, Qureshi2021NeRPNR}, in this work, we tackle the challenging problem of how one might automate this procedure, to enable a robot in a previously unseen home environment to determine a tidy configuration without any human intervention.

Human-inhabited environments such as homes and offices are designed to be functional and aesthetically pleasing. A key characteristic of such environments is the semantic organization, i.e., objects are placed in locations based on their purpose. This enables humans to adapt efficiently to new environments designed to serve the same purpose. For example, when a person enters a new home and wants to find sugar to make a cup of coffee, they instinctively look in the kitchen or pantry. %Conversely, when organizing a house, placing objects in a location that facilitates easy retrieval is a priority.
We leverage this semantic organization to enable robots to predict the likely locations of any given object. Specifically, we leverage recent developments in multimodal (vision-language) representation learning to propose a flexible approach for learning \textit{object-room affinities}, i.e., the relative likelihood of any given object belonging to a particular room in a house, based on image and text input.

% \textcolor{blue}{How do humans do it?/Source of Inspiration}
% Human Beings are complex. Over the course of the development of society, norms of how we arrange our houses have changed. Still, there are traces of common associations that we as the Human race have managed to stick to. 
% When a human being is tasked to find us "Headphones" they would intuitively look for it in places such as "Home Office", "Living Room", and "Entertainment Room". This small thought experiment gives us a good idea that subconsciously we have generated some associations between the objects and the rooms they are supposed to be in. For some objects this mapping is guided by "utility", for example: if asked to find "an apple" in an unknown house, you would first look for Kitchen, Pantry, or Dining Room. Whereas, for some objects, these mappings can be guided by "ease of use" for example: even though "utility" wise "water bottle" is supposed to be in and around the "kitchen", through experience, w know that we are more likely to find it in places where Humans are more prone to spend time.

% \textcolor{blue}{}
% Recently a lot of work has been done on generating a contextual vision and language embeddings. \textcolor{green}{something in praise of CLIP}
\begin{figure}[tb] 
        \centering
        \includegraphics[width=0.45\textwidth]{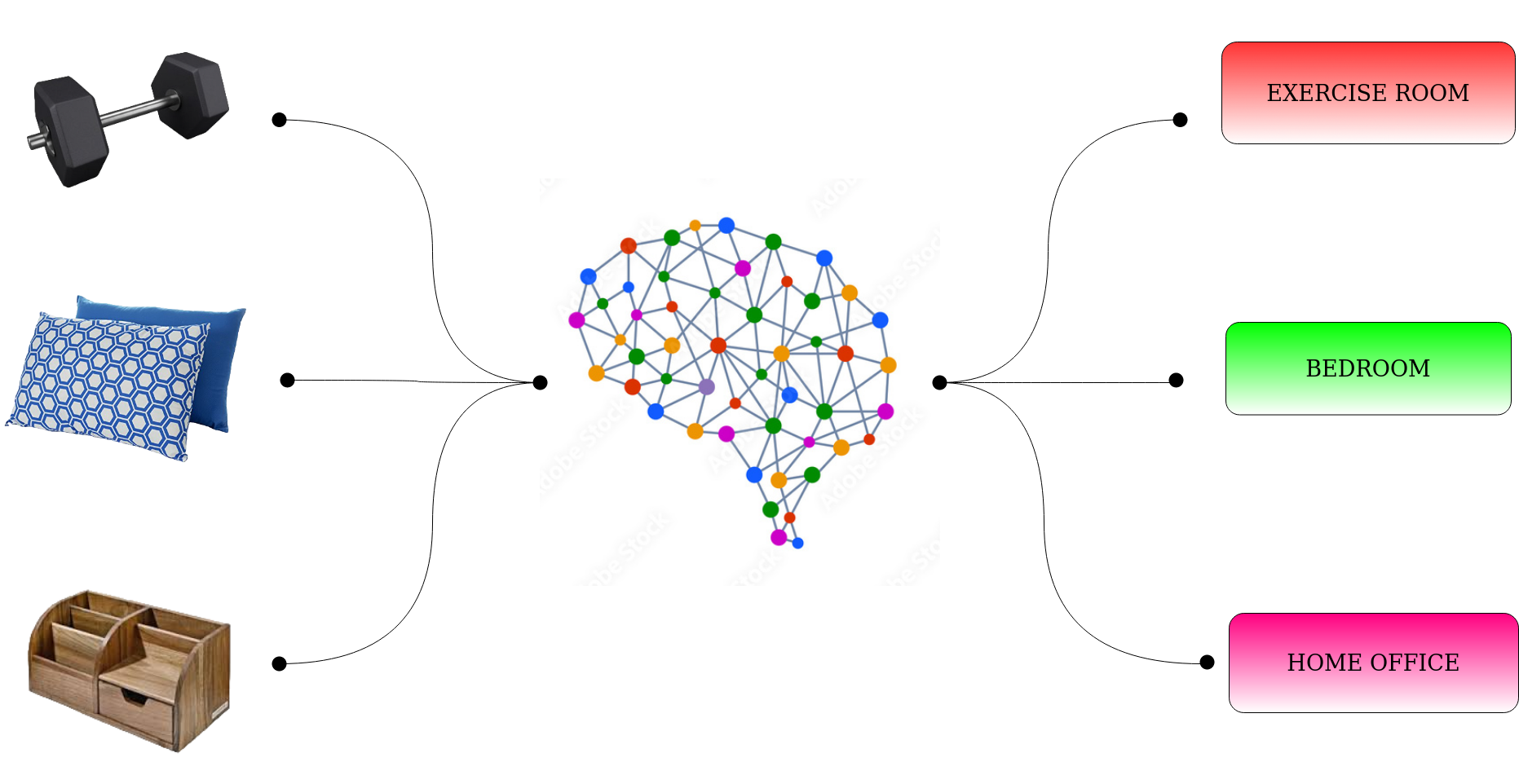}
        % \vspace{-1em}
        \caption{Our method leverages semantic organization (e.g., ``dumbbells are usually in the exercise room") to better compute the most suitable location for any given object.}
        % \href{https://drive.google.com/file/d/1Tt-0rLaSdmhIz-_ZKRvkYm5zGTHxOcZj/view?usp=sharing}{link}
        \label{figure:intro-fig}
        \vspace{-1em}
\end{figure}

State-of-the-art methods have used Large Language Models (LLMs) as \emph{commonsense} reasoning machinery for this \emph{tidy up} task~\cite{Kant2022HousekeepTV}.
These methods are limited to textual descriptors, which can be challenging to ground to a specific scene. Moreover, they use ground truth object labels for generating object-room affinities, which limits their operation outside of the training data distribution.
% and require accurate image segmentation, which is impractical in complex domains.
Others have used reinforcement learning (RL) to compute policies for related tasks such as visual semantic navigation~\cite{Chaplot2020ObjectGN, Majumdar2022ZSONZO, Gireesh2022ObjectGN,Gadre2022CLIPOW}, and Multi-Object Navigation~\cite{AyushMultiON, Ellis2023NavigationAM, Marza2021TeachingAH}, but do not fully leverage knowledge from different sources in the learning process.%multimodal learning or the general-purpose capabilities of modern language models.

Our framework, \emph{CLIPGraphs}, seeks to leverage the complementary strengths of commonsense knowledge, data-driven methods, and multimodal embeddings to estimate object-room affinities accurately. It does so by incorporating:
\begin{enumerate}
    \item A \emph{knowledge graph} that encodes human preferences of the room location of objects in home environments;
    
    \item Joint embeddings of image and text features~\cite{OpenCLIP} to support multimodal learning and queries in the form of images or text; and 
    
    \item A graph network that learns object-room affinities over a dataset of common household objects based on latent embeddings of the knowledge graph that includes the image and text feature embeddings. 
\end{enumerate}
The novelty is in the combination of these components to achieve the desired objective. We evaluate our framework's ability to correctly estimate the best room location for any given object, the key step in scene rearrangement. We do so using a dataset of 8000 image-text pairs that we created by extracting images from the Web for 268 benchmark object categories~\cite{Kant2022HousekeepTV}. We show experimentally that our framework substantially improves performance compared with state of the art baselines comprising LLMs and language embeddings encoding commonsense knowledge of the location of objects. 

\section{Related Work}
We motivate our novel framework by reviewing the limitations of related work.

\textbf{Embodied AI}:
To train embodied agents to perform human-like activities, many common tasks have being explored recently like goal navigation~\cite{Wijmans2019DDPPOLN, Anderson2018OnEO, Kim2022TopologicalSG, Kwon2021ImageGoalNA}, object navigation~\cite{Batra2020ObjectNavRO, AyushMultiON, Gireesh2022ObjectGN, Chaplot2020ObjectGN, Wortsman2018LearningTL, Yang2018VisualSN}, scene exploration~\cite{Chaplot2020LearningTE, Chaplot2020SemanticCF}, embodied QA~\cite{Das2017EmbodiedQA, Gordon2017IQAVQ, Cangea2019VideoNavQABT}, and rearrangement~\cite{batra2020rearrangement, Weihs2021VisualRR, trabucco2022simple}. 
ALFRED~\cite{Shridhar2019ALFREDAB}, TEACh~\cite{Padmakumar2021TEAChTE}, and \cite{Anderson2017VisionandLanguageNI} study the ability of agents to perform actions based on natural language instructions, and \cite{Chen2018IterativeVR, Marino2016TheMY, Wang2018ZeroShotRV} use knowledge graphs for visual classification and detection.
While these works include explicit specification of the goal state by a human agent, recent works have started the inclusion of reasoning with commonsense knowledge to enable agents to perform these tasks intelligently. 

\textbf{Commonsense Reasoning}
 In the context of rearrangement, Housekeep~\cite{Kant2022HousekeepTV}, and TIDEE~\cite{TIDEE} work on tidying a house using commonsense reasoning based on the training of Large Language Models (LLMs); and CSR~\cite{Gadre2022ContinuousSR} generates reasoning from a scene graph to detect objects and changes in room states. 
Other works like JARVIS \cite{Zheng2022JARVISAN}, DANLI \cite{Zhang2022DANLIDA}, and LLM-Planner \cite{song2023llmplanner} show the effectiveness of prompting LLMs for language understanding and sub-goal planning using natural language instructions.
\cite{Bisk2019PIQARA} evaluates the performance of different language models and studies their limitations concerning commonsense in the physical world.

 \textbf{CLIP for Embodied AI}
CLIP (Contrastive Language-Image Pre-training)~\cite{CLIP} uses large-scale text-image pairs for training image and text encoders simultaneously and has shown remarkable performance for object recognition.
The effectiveness of CLIP image and text embeddings for Embodied AI tasks has been evidenced by recent studies \cite{khandelwal2022simple, Shen2021HowMC, Weihs2021VisualRR, conceptfusion} over traditional ResNet-based architectures \cite{wijmans2020train}.
\cite{goodwin2021semantically} demonstrated the use of CLIP to match objects in a cross-instance setting with visual features as a measure of similarity to complete tabletop object rearrangement tasks. 
A recent work, ZSON~\cite{Majumdar2022ZSONZO}, proposes a zero-shot object navigation agent that uses CLIP embeddings to localize objects in the environment and navigate towards them without any additional training. The agent leverages the semantic similarity between the object category name and the visual features of the object to guide its exploration.
Similarly, CLIP was used by \cite{Dorbala2022CLIPNavUC, Dorbala2023CanAE} for zero-shot vision and language navigation by using natural language expressions for descriptions of target objects.
Recent works \cite{Huang2022VisualLM, Shah2022LMNavRN, Gadre2022CLIPOW, Chen2022OpenvocabularyQS, jain2022ground} use pixel-level CLIP features for robotic navigation using language commands.
\cite{Thomason2021LanguageGW, Shridhar2021CLIPortWA} have demonstrated the use of CLIP visual and language embeddings for learning robotic scenes, and \cite{Ha2022SemanticAO, CLIP-Fields} use CLIP for generating 3D scene memories from 2D images and natural language.
%%%%%%%%%%%%%%%%%%%%%%%%%%%%%%%%%%%%%%%%%%%%%%%%%%%%%%%%%%%%%%%%%%%%%
%%%%%%%%%%%%%%%%%%%%%%%%%%%%%%%%%%%%%%%%%%%%%%%%%%%%%%%%%%%%%%%%%%%%%

%%%%%%%%%%%%%%%%%%%%%%%%%%%%%%%%%%%%%%%%%%%%%%%%%%%%%%%%%%%%%%%%%%%%%
%%%%%%%%%%%%%%%%%%%%%%%%%%%%%%%%%%%%%%%%%%%%%%%%%%%%%%%%%%%%%%%%%%%%%
\section{Problem Formulation and Framework}

% \subsection{Task Description}
To perform tidying up or other scene rearrangement tasks, a robot needs the key ability to accurately compute the appropriate location for any given object. % For introducing commonsense reasoning in tasks like navigation and rearrangement, the agent needs to identify objects and their expected correct locations in a \emph{tidy} house. 
To explore this ability, we created the \textit{Images for Room-Object Nexus through Annotations} (IRONA) dataset of $30$ RGB images from the Web for each of the $268$ categories of household objects used by Housekeep~\cite{Kant2022HousekeepTV}\footnote{Supplementary material at: \href{https://clipgraphs.github.io}{https://clipgraphs.github.io}\label{supplementary-material}}. For any such image, the robot had to compute the likelihood that the object in the image belongs to each of $17$ room categories.

\begin{figure*}[tb] 
        \centering
        % \includesvg[inkscapelatex=false,width=\textwidth]{sections/figures/Intoduction_Figure.drawio.svg}        
        \includegraphics[width=0.75\textwidth]{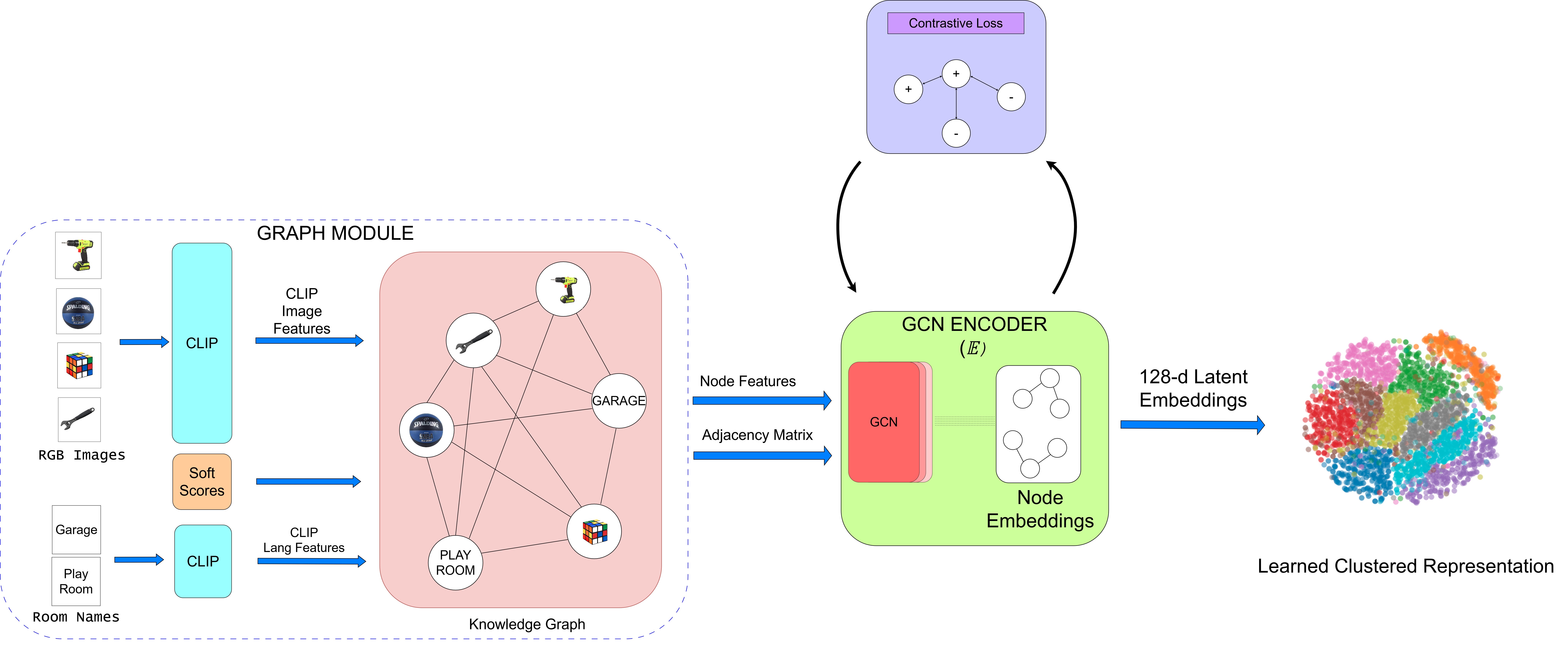}
        % \vspace{-1em}
        \caption{\emph{CLIPGraphs} constructs a graph module (bottom-left) using CLIP encoders and passes that to a GCN Encoder(\emph{$\mathbb{E}$}) module. The encoder is trained using contrastive loss to create better node embeddings that bring similar embeddings closer. Visualization of  final layer activations confirms the formation of well-defined node clusters.}
        \label{figure:pipeline}
        % \vspace{-1em}
\end{figure*}

Our framework, called CLIPGraphs, trains a Graph Convolutional Network (GCN)~\cite{kipf2017semi} to compute embeddings that are used to estimate these object-room affinities. Figure~\ref{figure:pipeline} shows the training pipeline. It uses a knowledge graph to encode existing information of human preferences (of room location of objects) for the object categories~\cite{Kant2022HousekeepTV}, and incorporates a modified contrastive loss function to compute better latent embeddings of the image and language encoder features provided by CLIP~\cite{CLIP} for the nodes of the knowledge graph. The resultant node embeddings model the information about the room location of various objects in the latent space. During inference, the CLIP features generated for any (test) RGB images are processed by the GCN, with the cosine similarity between the embeddings of the rooms and the image providing the desired estimate of object-room affinities. We describe individual components of our framework below.

%%%%%%%%%%%%%%%%%%%%%%%%%%%%%%%%%%%%%%%%%%%%%%%%%%%%%%%%%%%%%%%%%%%%%
\subsection{Knowledge Graph}\label{section:knowledge-graph}
% Our graph is over 268 objects and 17 room categories provided by Housekeep \cite{Kant2022HousekeepTV}. 
Our framework's first step uses the human-annotated preferences included in the Housekeep data~\cite{Kant2022HousekeepTV}. For every object-room pair, 10 human annotators ranked the receptacles in that room based on the likelihood of the object being placed there correctly or incorrectly. For each object-room-receptacle tuple, there are thus 10 opinions that could be positive, negative, or zero. We filter the dataset to ensure good agreement amongst annotators. We calculate the positive (negative) soft scores as the mean of the positive (negative) reciprocal preference of all the receptacles for a given object-room pair. To establish ground truth object-room mappings, we use the object-room-receptacle scores, i.e., we select the room with the highest positive-scored receptacle. Every other room in the domain is assigned the mean negative soft score of receptacles in that room\footref{supplementary-material}.

\begin{figure}[tb] 
        \centering     
        \includegraphics[width=0.45\textwidth]{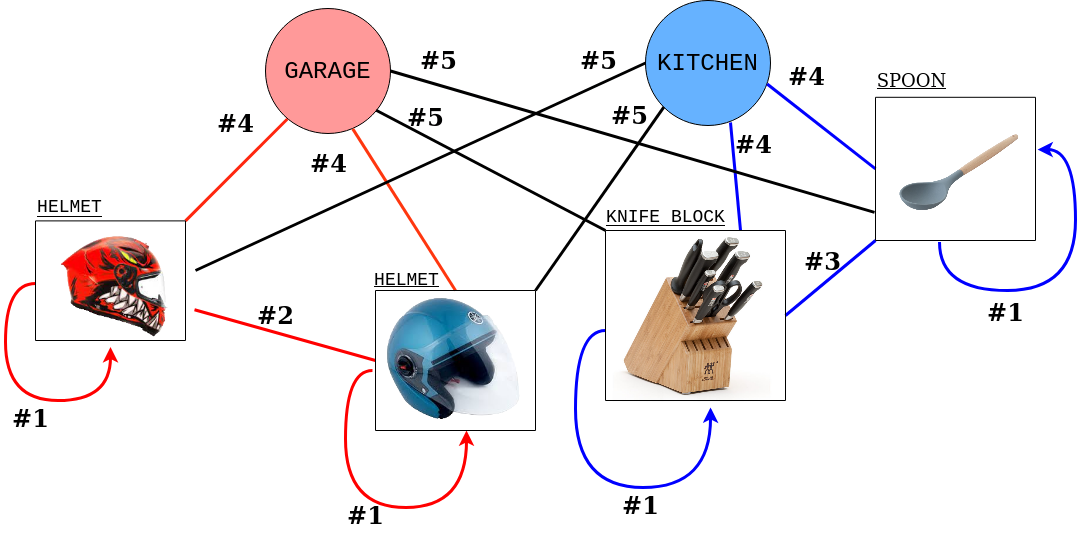}
        \caption{An illustration of the five types of edges in our knowledge graph. The colored edges denote positive edge weights whereas black ones denote negative weights. The number on the edge denotes the type of edge.}
        \vspace{-1em}
        \label{figure:edges}
\end{figure}

To use the available annotated information to populate a knowledge graph, we partitioned the IRONA web-scraped dataset into training, validation, and test sets in a ratio of 15:5:10 images per object category. The knowledge graph is instantiated with each image of the training set as a node, along with room names, i.e., there are 268*15 + 17 = 4037 nodes. We then considered five types of edges connecting nodes (see Figure~\ref{figure:edges}): (1) image self edge (edge weight=1); (2) edge between images of same object (edge weight=1); (3) edge between two objects in the  same ground truth room; (4) edge between object and its correct room node; and (5) edge between object and its incorrect room nodes. Next, we assigned weights for each type of edge. Weights for edges of type 4 and 5 were based on the object-room soft scores. Edges of type 3 were given a randomly chosen weight between $0.5$ to $0.7$, and edges of type 1 and 2 were assigned a weight of $1$.
% \end{enumerate}

Once the basic knowledge graph is created, we initialize the graph's nodes using the pretrained CLIP model's high-dimensional embeddings. Specifically, each object node is initialized with the corresponding CLIP image encoder embedding, and each room node is initialized with the corresponding CLIP language encoder embedding. This is because we want to capture the appearance of the objects and the known association between objects and rooms (based on the large dataset used to train CLIP embeddings). In particular, we considered three pretrained architectures of CLIP in our experiments: Vision Transformer (ViT), ResNet-50, and ConvNeXt. ViT-H/14~\cite{vit_h_14} is trained on LAION-2B, which is a 2.3 billion subset of the LAION-5B~\cite{laion5b} dataset with English captions. ResNet-50~\cite{RN50} uses OpenAI's pretrained weights~\cite{CLIP}, and ConvNeXt base~\cite{convnet} is pre-trained on LAION-400m~\cite{laion400m}, which contains 400 million image-text pairs\footnote{Implementation used existing code~\cite{ilharco_gabriel_2021_5143773}.}. For a discussion about how we experimentally chose the embedding for different nodes, please refer to our supplementary material. Once we have associated CLIP embeddings with our knowledge graph's nodes, we move to the next steps of our training pipeline.

 \subsection{GCN Training}
The next step of training feeds these node embeddings, each of 512 or 1024 dimensions based on the CLIP architecture chosen, and the adjacency matrix (of knowledge graph structure) to a Graph Convolutional Network (GCN)~\cite{kipf2017semi} to learn better latent space embeddings of our knowledge graph. GCNs are able to capture non-linear relationships between nodes, and learn from both local and global structures in a graph. As a result, nodes that are more similar are mapped to points that are closer in the latent embeddings space, whereas nodes that are dissimilar are mapped to points further away in the latent space. For example, the output 128-dimensional GCN (object) embedding for a microwave will have a higher cosine similarity with the output 128-dimensional GCN (language) embedding for the kitchen. 

An important design decision during training is the choice of the loss function. Prior work has devoted much attention to functions such as contrastive loss~\cite{Sermanet2017TimeContrastiveNS}, triplet loss~\cite{Schroff2015FaceNetAU}, and multi-class N-pair loss~\cite{sohn2016improved}. Recent work has demonstrated the benefits of using the loss function introduced in the CLIP-Fields method~\cite{CLIP-Fields}. We modify this loss function to further leverage the knowledge graph created using the IRONA dataset and human preference annotations.

\begin{figure}[tb] 
        \centering
        \includegraphics[width=0.3\textwidth]{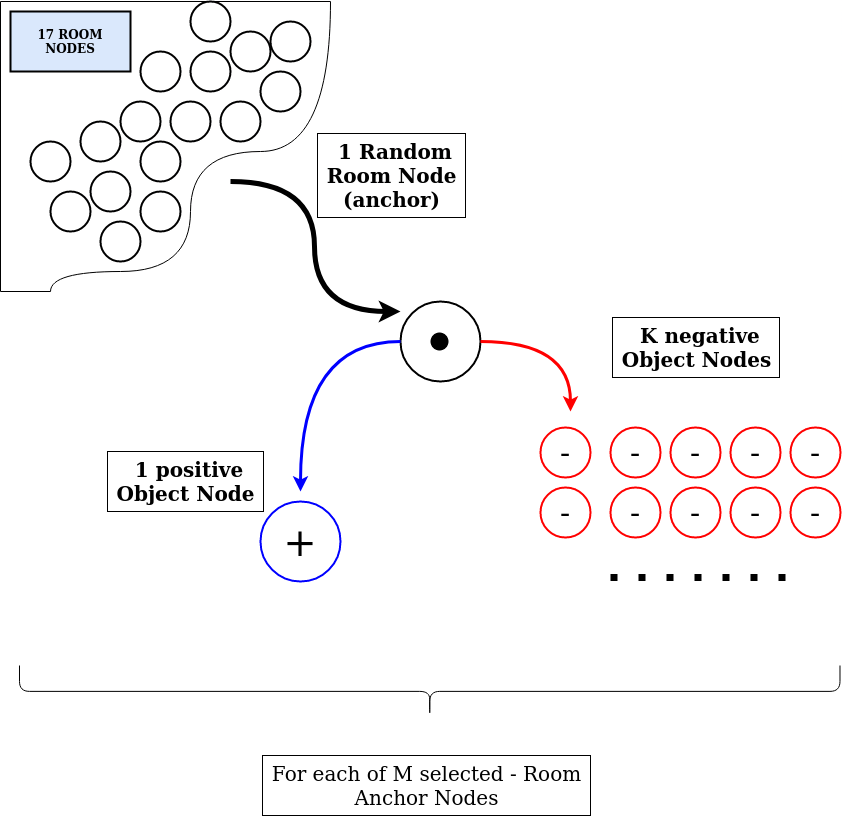}
        % \vspace{-1em}
        \caption{Sampling method used in the loss function; shown for $K$ = 10 and $M$ = 1; we average the loss over $M$ batches.}
        \vspace{-1em}
    \label{fig:sample-2a}
\end{figure}

 %%%%%%%------------------------------------------
\noindent
\textbf{Loss Function.}
% Previously contrastive losses\cite[],triplet loss \cite{} have been explored well to learn clustered latent embeddings.
% This loss function aims at improving the connection between the anchor and the positive sample while increasing the distance with the negative sample. 
We train our GCN using a contrastive loss function similar to that described in the CLIP-Fields method~\cite{CLIP-Fields} with the objective of clustering similar embeddings closer in the latent space and mapping dissimilar embeddings to points that are further away in the latent space. We adapt the basic loss function to our problem formulation and use the additional information of edge weights.
\begin{equation}\label{eq:con_loss}
L = - e^{-weight_{{\textbf{+}\bullet}}} \log \left( \frac{e^{({sim_{\textbf{+}\bullet} / T)}}}{\sum_{i=1}^{K} e^{({sim_{\textbf{--}\bullet , i} / T)}}} \right)
\end{equation}
where $weight_{+.}$ is the edge weight  between the positive node and the anchor node, $sim_{\textbf{+}\bullet}$ is the cosine similarity between the anchor and a positive node embedding, and $sim_{\textbf{--}\bullet, i}$ is the cosine similarity between anchor node embedding and $i^{th}$ negative node embedding. $T$ is a temperature term that is tuned over a validation set.
% We elaborate more upon this and discuss the sampling method in \ref{train}.
%
We randomly select one of the 17 rooms as our anchor node, then choose a positive node (for numerator in Equation~\ref{eq:con_loss}) by picking an object within that room at random, and finally sample $k$ negative nodes for the denominator of the loss function from objects located outside the room; Figure~\ref{fig:sample-2a} illustrates this process, which is repeated for a batch of samples and the mean loss is calculated. This formulation of the loss function minimizes the distance between the anchor node and the positive node while maximizing the distance with each of the negative nodes, leading to distinct clusters in the graph embeddings. As stated before, the training pipeline is outlined in Figure~\ref{figure:pipeline}.
%
% This sort of embedding generation is further improved during training using contrastive loss functions which encourage similar nodes in a network to be close together while pushing dissimilar ones further apart in an embedded space. 
% This leads to the formation of distinct clusters in the graph embeddings.

% However, as there are multiple classes in the dataset, it has been shown \cite{sohn2016improved} that choosing multiple negative samples per anchor point increases the distance between multiple classes at once. \textbf{\textcolor{red}{should we add sth from loss ablations part?}} This helps reduce noise by encouraging distinct clusters within the dataset.

%% Add more references

\begin{figure*}[h!] 
        \centering
        \includegraphics[width = 0.65\textwidth]{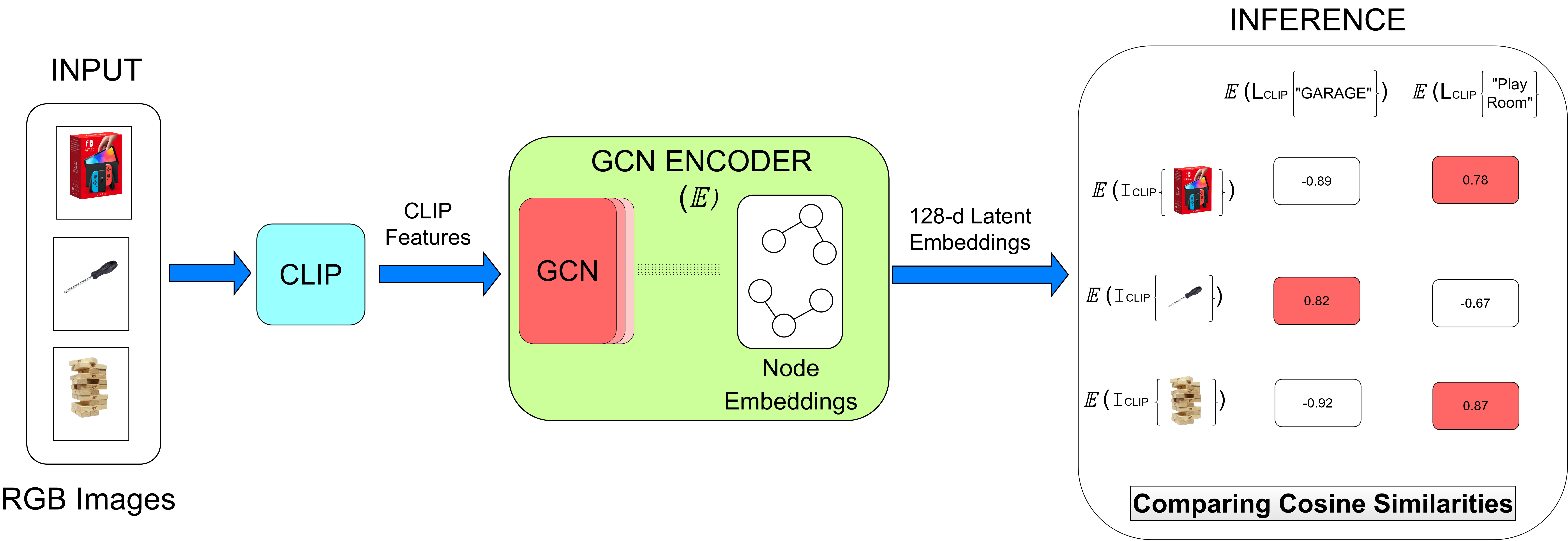}
        % \vspace{-1em}
        \caption{Our inference pipeline processes input RGB images to generate CLIP image embeddings. These embeddings are processed by the GCN Encoder to produce latent image embeddings. Cosine similarity between these latent embeddings and previously learned room embeddings determines object-room affinities.}
        \label{fig:test_pipeline}
        % \vspace{-1em}
\end{figure*}

%
% As stated before, the  training pipeine is outlined in Figure~\ref{figure:pipeline} shows the general representation of our training process.
% We make use of graph convolutional operator from \cite{kipf2017semi}
%
% We divide the IRONA Dataset's 8040 images of the 268 object categories into train, test and validation sets in 15:10:5 proportion, and generate their initial embeddings using the visual encoders of CLIP.
% %
% We further add room embeddings for the 17 rooms to the graph using the language encoder of the equivalent CLIP architecture.
% %
%
% \begin{equation}\label{eqn:gcn}
% \mathbf{x}^{\prime}_i = \mathbf{\hat{D}}^{-1/2} \mathbf{\hat{A}} \mathbf{\hat{D}}^{-1/2} \mathbf{x}_i \mathbf{\Theta}    
% \end{equation}

% where $\mathbf{x}_i$ and $\mathbf{x}^{\prime}_i$ are the input and output node features of node $i$, respectively. $\mathbf{\hat{A}}$ denotes the adjacency matrix with inserted self-loops and $\mathbf{\hat{D}}$ is its diagonal degree matrix. $\mathbf{\Theta}$ denotes a learnable weight matrix.

\subsection{Testing}\label{test} 
Once the GCN has been trained, the pipeline used for testing (i.e., inference) is shown in Figure~\ref{fig:test_pipeline}. Similar to the process of training, we compute the CLIP image encoder embedding for the test image, and the CLIP language encoder embedding for the possible rooms. These embeddings are passed to the GCN with only self-edges (in the absence of a knowledge graph) to obtain the output (latent space) embedding for the test image and the possible rooms. Next, similarity scores are calculated between each image node $\vec{x}$ and each of the room(s) $\vec{y}$ using the cosine similarity function:
\(
\cos(\vec{x},\vec{y}) = \frac{\vec{x} \cdot \vec{y}}{||\vec{x}|| \cdot ||\vec{y}||}   
\).
We then average the similarity scores over different images of each object category to get the affinity score between that object category and each of the candidate rooms.

%%%%%%%%%%%%%%%%%%%%%%%%%%%%%%%%%%%%%%%%%%%%%%%%%%%%%%%%%%%%%%%%%%%%%
%%%%%%%%%%%%%%%%%%%%%%%%%%%%%%%%%%%%%%%%%%%%%%%%%%%%%%%%%%%%%%%%%%%%%
\section{Experimental Setup and Results}
This section describes the experiments we conducted and discusses the corresponding results.

%%%%%%%%%%%%%%%%%%%%%%%%%%%%%%%%%%%%%%%%%%%%%%%%%%%%%%%%%%%%%%%%%%%%%
\subsection{Experimental Setup}
Object-room affinities have predominantly been determined by language-based embeddings or human input in prior work. Since our work combines prior knowledge and  multimodal (vision, language) inputs, our chosen baselines were off-shelf language encoders and the GPT-3 LLM. %For each language model baseline, we generate the cosine similarities between the object name and the room name to rank all 17 rooms in decreasing order of their affinities. 
We experimentally evaluated the following hypothesis:
\begin{itemize}
    \item \textbf{H1:} CLIP language embeddings result in better performance than other language encoder embeddings;
    
    % Our framework provides better performance than the language encoder embeddings and the GPT-3 LLM. 
    \item \textbf{H2:} Multimodal CLIP embeddings, by themselves, do not perform better than language-based embeddings; % Our multimodal embedding results in better performance than just the vision or language embeddings. %CLIP Embeddings have a lot of global contexts embedded within them. Their Image Encoder embeddings don’t just express visual features but have an aspect of language common sense as well. This makes them easily fine-tuneable. In the battle of language vs vision features, a multimodal representation like that of CLIP emerges as a clear winner.
    % \item \textbf{H3:} Our framework outperforms the GPT-3 LLM.
    
    \item \textbf{H3:} Our framework leads to better performance than (i) the underlying CLIP embedding, (ii) just the language-based encodings, and (iii) the GPT-3 LLM; %As compared to unfinetuned CLIP Features, our GCN-based fine-tuning leads to a better latent embedding clustering that translates to a better common sense understanding of object-room relationships  
    \item \textbf{H4:} Our framework provides robustness to previously unseen noisy backgrounds. % on clean images  works well for noisy and real-world images
\end{itemize}
We evaluated \textbf{H1-H3} quantitatively and evaluated \textbf{H4} qualitatively. The performance task was to compute estimates of object-room affinities for all 268 object categories and 17 rooms in the test split of the IRONA dataset.
% We considered 2 baselines for evaluation:
% \begin{enumerate}
%     \item \textbf{Language Encoders:} We take off the shelf un-finetuned Language encoders to compare how efficiently do they express \emph{object-room} connection in their latent space embeddings
%     \item \textbf{GPT-3:} We task a powerful LLM like GPT-3 with the same task, and ask it to rank all the 17 rooms for all 268 object categories.  
% \end{enumerate}
We considered two performance measures:
\begin{enumerate}
    \item \textbf{mAP:} The mean average precision (mAP) is the average of precision scores at different recall values for each instance of an object category, and the mean over all the object categories. %We use the implementation by \cite{scikit-learn}
    \item \textbf{Top $k$ Hit Ratio:} The average fraction of object categories for which the ground truth correct room was among the Top $k$  estimates from our framework.
\end{enumerate}
All claims are statistically significant unless stated otherwise.

%%%%%%%%%%%%%%%%%%%%%%%%%%%%%%%%%%%%%%%%%%%%%%%%%%%%%%%%%%%%%%%%%%%%%
\subsection{Quantitative Results}
To evaluate \textbf{H1}, we first compared two existing language encoder embeddings (RoBerta~\cite{roberta}, GloVE~\cite{glove}) with just the CLIP-based language embeddings with each of the three CLIP architectures. As shown in Table~\ref{table:lang_baseline}, the CLIP-based language embeddings (particularly the ViT architecture) resulted in better performance, supporting \textbf{H1}. 
%
% For this, we generate the language embeddings of every object category and room and compute the similarity scores between them to calculate the metrics. 
%
% \\

Next, we compared the performance of the multimodal (vision, language) CLIP embeddings for each of the three CLIP architectures. As shown in Table~\ref{table:CLIP-untrained}, performance is comparable but slightly worse than that in Table~\ref{table:lang_baseline}.  These results support \textbf{H2} and motivate the use of GCNs.

% Since CLIP provides both language and image encoders, we also make use of the visual encoders to generate the embedding for each object, and the language encoder to generate the embedding for each room. 
%

% Without any further finetuning of the model, if we take the cosine similarity of the CLIP embeddings for images and rooms, the results are highlighted in Table \ref{table:CLIP-untrained}.
%
% We observe that in both of these cases, without any fine-tuning of the models the test mAP \approx 0.42 - 0.45

\begin{table}[tb]
    \vspace{-1em}
    \centering
    \setlength\tabcolsep{11pt}
    \begin{tabular}{lSSSS}
        \toprule
        Lang Model & {Test mAP $\Uparrow$} & \multicolumn{3}{c}{Hit-Ratio $\Uparrow$} \\
        \cmidrule{3-5}
         & & {Top-1} & {Top-3} & {Top-5}\\
        \midrule
        ConvNeXt & {0.405} & {0.223} & {0.472} & {0.632} \\
        ViT & \textbf{0.456} & {0.256} & \textbf{0.576} & \textbf{0.710} \\
        RN50 & {0.453} & \textbf{0.275} & {0.546} & {0.643} \\
        RoBerta & {0.417} & {0.238} & {0.491} & {0.636} \\
        GloVE & {0.148} & {0.123} & {0.208} & {0.278} \\
        \bottomrule
    \end{tabular}
    \caption{CLIP-based language embeddings perform better than other popular language encoders; results support \textbf{H1}.}
    \label{table:lang_baseline}
    \vspace{-1em}
\end{table}

\begin{table}[tb]
    \centering
    \setlength\tabcolsep{10pt}
    \begin{tabular}{lSSSS}
        \toprule
        UnTuned-CLIP & {Test mAP $\Uparrow$} & \multicolumn{3}{c}{Hit-Ratio $\Uparrow$} \\
        \cmidrule{3-5}
         & & {Top-1} & {Top-3} & {Top-5}\\
        \midrule
        ConvNeXt & {0.41} & {0.24} & {0.46} & {0.62} \\
        ViT & \textbf{0.42} & \textbf{0.25} & \textbf{0.49} & \textbf{0.65} \\
        RN50 & {0.39} & {0.19} & {0.45} & {0.67} \\
        \bottomrule
    \end{tabular}
    \caption{Multimodal CLIP embeddings, by themselves, do not improve performance compared with just the CLIP-based language embeddings (see Table~\ref{table:lang_baseline}). Results support \textbf{H2}.} %Table showing results for embeddings without GCN based fine tuning \textbf{supports H4}; Zero Shot Prediction [To highlight the need for GCN based tuning This table shows a comparison of various CLIP models without fine-tuning]}
    \label{table:CLIP-untrained}
    \vspace{-1em}
\end{table}

Next, we computed the performance of our architecture, i.e., with GCNs trained using the contrastive loss function and the underlying multimodal CLIP embeddings, with the corresponding results shown in Table \ref{table:our-model}. The best performance was (once again) with the ViT version of the CLIP architecture. Also, performance was substantially better than with the multimodal CLIP embeddings (Table~\ref{table:CLIP-untrained}) or CLIP's language encoder embeddings (Table~\ref{table:lang_baseline}). For example, there is an $\approx 40\%$ increase in mAP score compared with not using the GCNs. These results partially support \textbf{H3}. 

% To improve on the information contained in CLIP embeddings, we extend them using a Graph Convolutional Network trained on the contrastive loss. 
% Keeping the same format of image encodings for the objects and language embeddings for the rooms, training the GCN improves the performance on CLIP as shown in Table \ref{table:our-model}. 
% We observe $\approx 40\%$ increase in the mean average precision score by fine-tuning using GCN.
% This experiment supports our hypothesis on the improvement of object-room affinities by building on CLIP embeddings.
\begin{table}[tb]
    \vspace{-1em}
    \centering
    \setlength\tabcolsep{11pt}
    \begin{tabular}{lSSSS}
        \toprule
        GCN-CLIP & {Test mAP $\Uparrow$} & \multicolumn{3}{c}{Hit-Ratio $\Uparrow$} \\
        \cmidrule{3-5}
         & & {Top-1} & {Top-3} & {Top-5}\\
        \midrule
        ConvNeXt & {0.73} & {0.62} & {0.81} & {0.88} \\
        ViT & \textbf{0.85} & \textbf{0.76} & \textbf{0.93} & \textbf{0.97} \\
        RN50 & {0.66} & {0.53} & {0.75} & {0.81} \\
        \bottomrule
    \end{tabular}
    \caption{CLIPGraphs use of GCN embeddings of multimodal CLIP features and commonsense knowledge results in substantially better performance compared with just the CLIP embeddings in Tables~\ref{table:lang_baseline} and~\ref{table:CLIP-untrained}. Results support \textbf{H3}.} % Result: This table compares shows a comparison of various CLIP models when fine-tuned using our GCN Model, using the loss function, sampling method, and hyperparameters obtained from the ablation studies. [\emph{see Supplementary material}]}
    %\vspace{0.4cm}
    \label{table:our-model}
\end{table}

To further explore the benefits of a multimodal CLIP representation, we conducted experiments with our framework,  but with GCN embeddings of only the language-based encoding of CLIP. The results reported in Table~\ref{table:lang-claim} show the benefits of using the multimodal CLIP embeddings. % which demonstrates the effectiveness of our model in cross-modal commonsense reasoning.

%%%%%SINCE WE ALSO CLAIM THAT CLIP EMBEDDINGS LIE IN AN EMBEDDING SPACE WHICH IS MORE LIKE A MULTIMODAL REPRESENTATION; WE VERIFY THIS HYPOTHESIS BY GIVING LANG EMBEDDINGS INSTEAD OF IMAGE EMBEDDINGS AT THE INFERENCE TIME AS WELL 
\begin{table}[tb]
    \vspace{-1em}
    \centering
    \setlength\tabcolsep{10pt}
    \begin{tabular}{lSSSS}
        \toprule
        GCN-CLIP[Lang] & {Test mAP $\Uparrow$} & \multicolumn{3}{c}{Hit-Ratio $\Uparrow$} \\
        \cmidrule{3-5}
         & & {Top-1} & {Top-3} & {Top-5}\\
        \midrule
        ConvNeXt & {0.64} & {0.53} & {0.69} & {0.76} \\
        ViT & \textbf{0.77} & \textbf{0.68} & \textbf{0.77} & \textbf{0.83} \\
        RN50 & {0.59} & {0.46} & {0.63} & {0.74} \\ 
        \bottomrule
    \end{tabular}
    \caption{Using our GCN-based embedding with just the underlying language-based CLIP encoding results in better performance than in the absence of the GCN embedding, but performance is not as good as when GCNs are used with the multimodal CLIP embeddings (in Table~\ref{table:our-model}).}%Claim verification [\textbf{H2}]: Since we also claim that CLIP embeddings are more multimodal representation than a normal visual feature; we verify this hypothesis by giving language embeddings instead of image encoder embeddings at the inference time.}
    \label{table:lang-claim}
    % \vspace{-1em}
\end{table}

The next experiment compared our framework's performance with the GPT-3 LLM and a state of the art language encoder that provided the best performance among language-based encoders. The results  %Next prepare another baseline by querying GPT-3 \cite{gpt-3} to generate rankings of the 17 rooms for each object and evaluate it's performance. 
summarized in Table~\ref{table:final-comparison} show that our framework provides substantially better performance by fully leveraging prior commonsense knowledge and multimodal CLIP embeddings. These results strongly support \textbf{H3}. %summarizes the results of the GCN model with two baselines: GPT-3 and language encoder.

\begin{table}[tb]

\begin{center}
    \setlength\tabcolsep{8pt}
    \begin{tabular}{@{}lSSSSS@{}}
    
    \toprule
    \multirow{3}{*}{} &
      \multicolumn{1}{c}{Test mAP $\Uparrow$} & 
      \multicolumn{3}{c}{Hit-Ratio $\Uparrow$} \\
      % \multicolumn{3}{c}{ }\\
      \cmidrule{2-3} \cmidrule{4-6} 
      & & {Top-1} & {Top-3} & {Top-5}\\
      \midrule
    Our[GCN-CLIP] [\ref{table:our-model}] & \textbf{0.85} & \textbf{0.76} & \textbf{0.93} & \textbf{0.97} \\
    GPT-3 & {0.66} & {0.52} & {0.76} & {0.81} &\\
    Best Lang Encoder[\ref{table:lang_baseline}] & {0.456}  & {0.275} & {0.576} & {0.71} \\
    \bottomrule
  \end{tabular}
    \caption{Our framework, with GCN and underlying multimodal CLIP embeddings, substantially improves performance compared with standalone GPT-3 LLM and language-based encoders; hence, the results strongly support \textbf{H3}.} %This is the main table;\textbf{supports H1,H3}  which compares best GCN fine-tuned CLIP model vs GPT-3 and the CLIP Unfinetuned Language Encoder}
    \vspace{-2em}
\label{table:final-comparison}

\end{center}
\end{table}

%%%%%%%%%%%%%%%%%%%%%%%%%%%%%%%%%%%%%%%%%%%%%%%%%%%%%%%%%%%%%%%%%%%%%
\subsection{Qualitative Results}
\begin{figure*}[tb] 
    % \vspace{em}
    \centering     
    \includegraphics[width =0.95\textwidth]{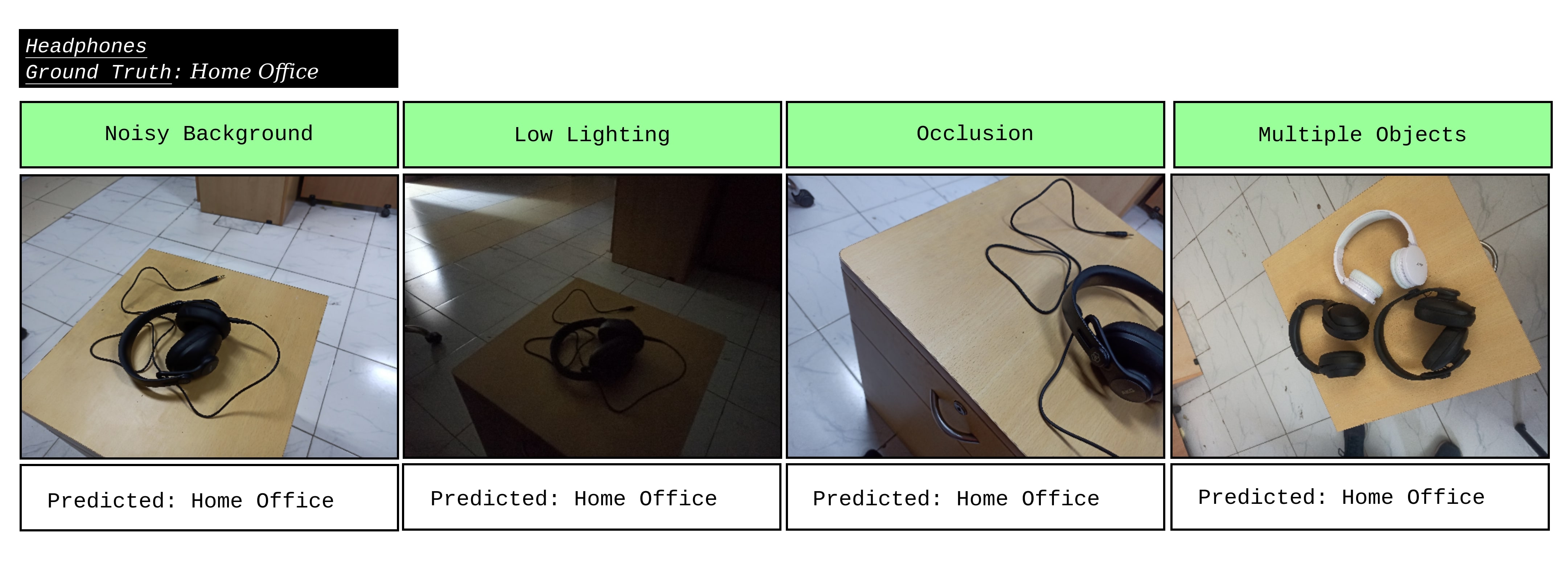}
    \caption{Qualitative results for previously seen objects in new backgrounds; supports \textbf{supports H4}.}
    \label{figure:qual-seen-success}
    % \vspace{1em}
\end{figure*}

\begin{figure}[!htbp] 
    \vspace{-1em}
    \centering     
    % \hspace{1pt}
    \includegraphics[width =0.35\textwidth]{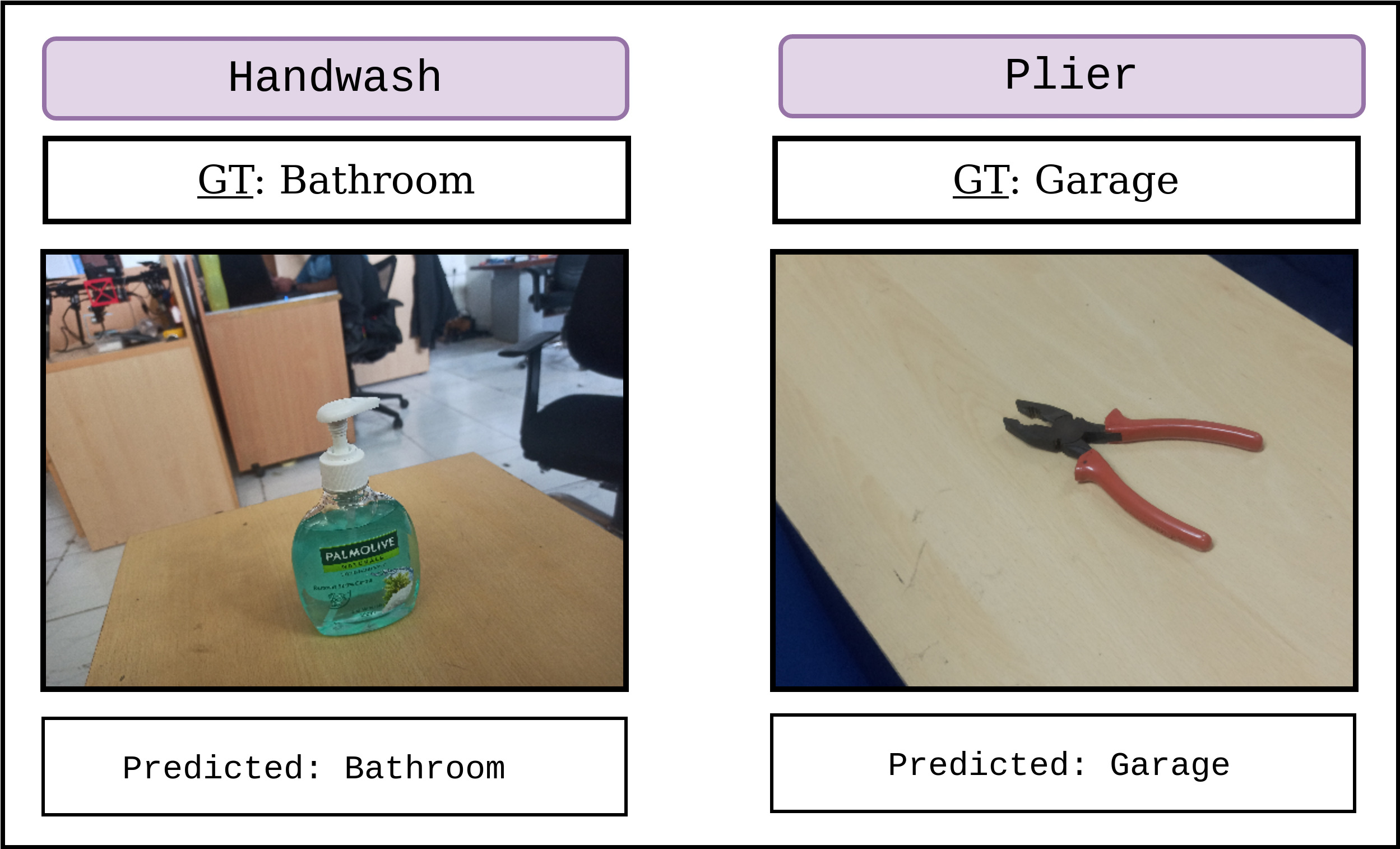}
    \caption{Successful placement of previously unseen object categories (Handwash, plier) in the correct room by leveraging commonsense domain knowledge.}%Example for Unseen Object Category; Handwash,Plier[not in our train set] was predicted correctly}
    \label{figure:qual-success:handwash}
    % \vspace{-1em}
\end{figure}

% \begin{figure}[tb]
%     \vspace{-1em}
%     \centering     
%     \includegraphics[width =0.4\textwidth]{sections/figures/failure-seen.png}
%     \caption{Failure to determine correct room location for \emph{power drill, book}, categories present in the training dataset.}
%     \label{figure:qual-fail:seen}
%     \vspace{1em}
% \end{figure}

\begin{figure}[!htbp]
    \vspace{-1em}
    \centering 
    \includegraphics[width =0.35\textwidth]{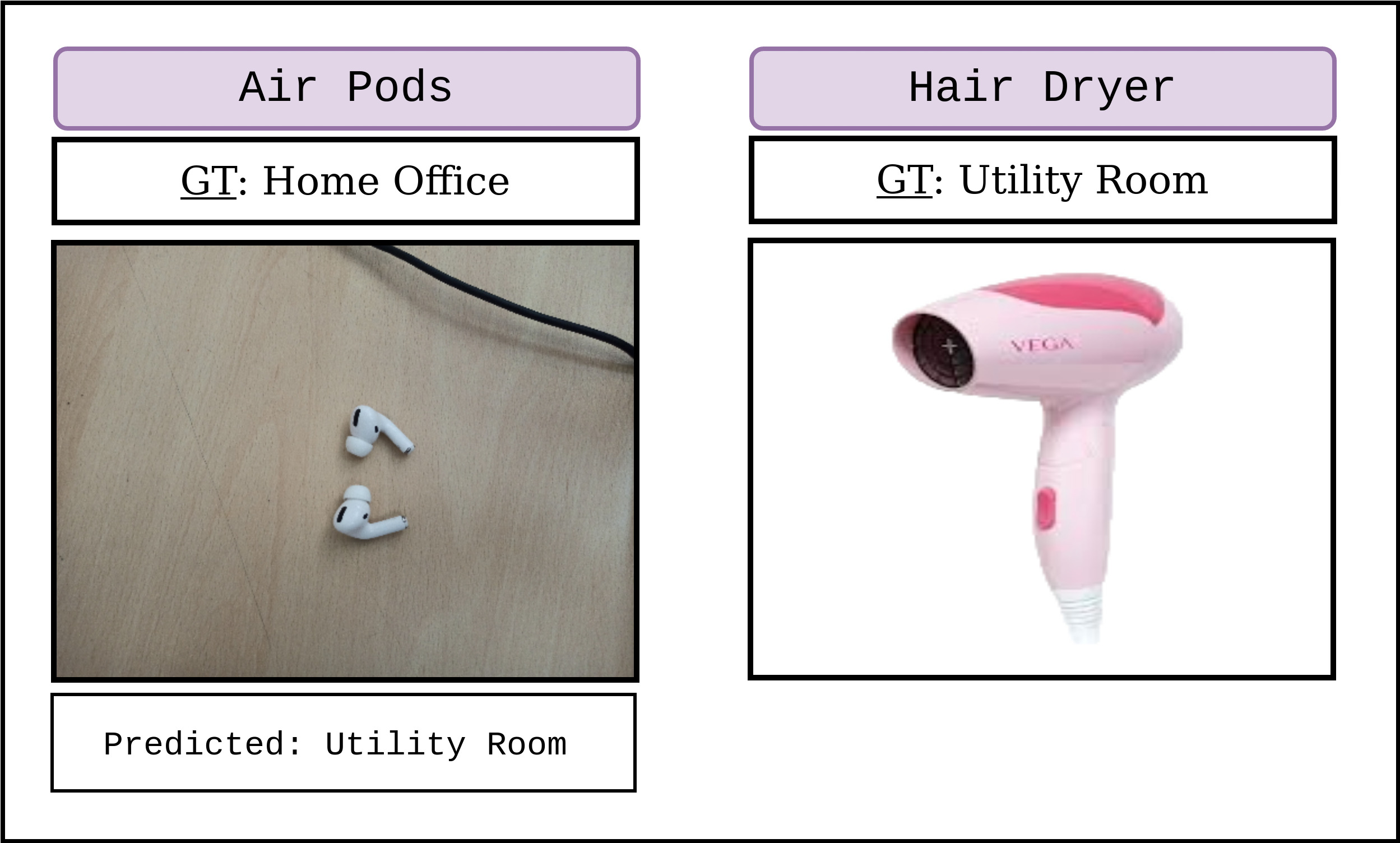}
    \caption{Failure to determine correct room for object category \textit{earpods} (not in our train set) because it was structurally similar to \textit{hair dryer} category that was in our training set.}%[in our train set] ; and thus predicts incorrectly\newline}
    \label{figure:qual-failure:1-air pods}
    % \vspace{-1em}
\end{figure}

\begin{figure}[!htbp]
    \vspace{-1em}
    \centering     
    \includegraphics[width =0.35\textwidth]{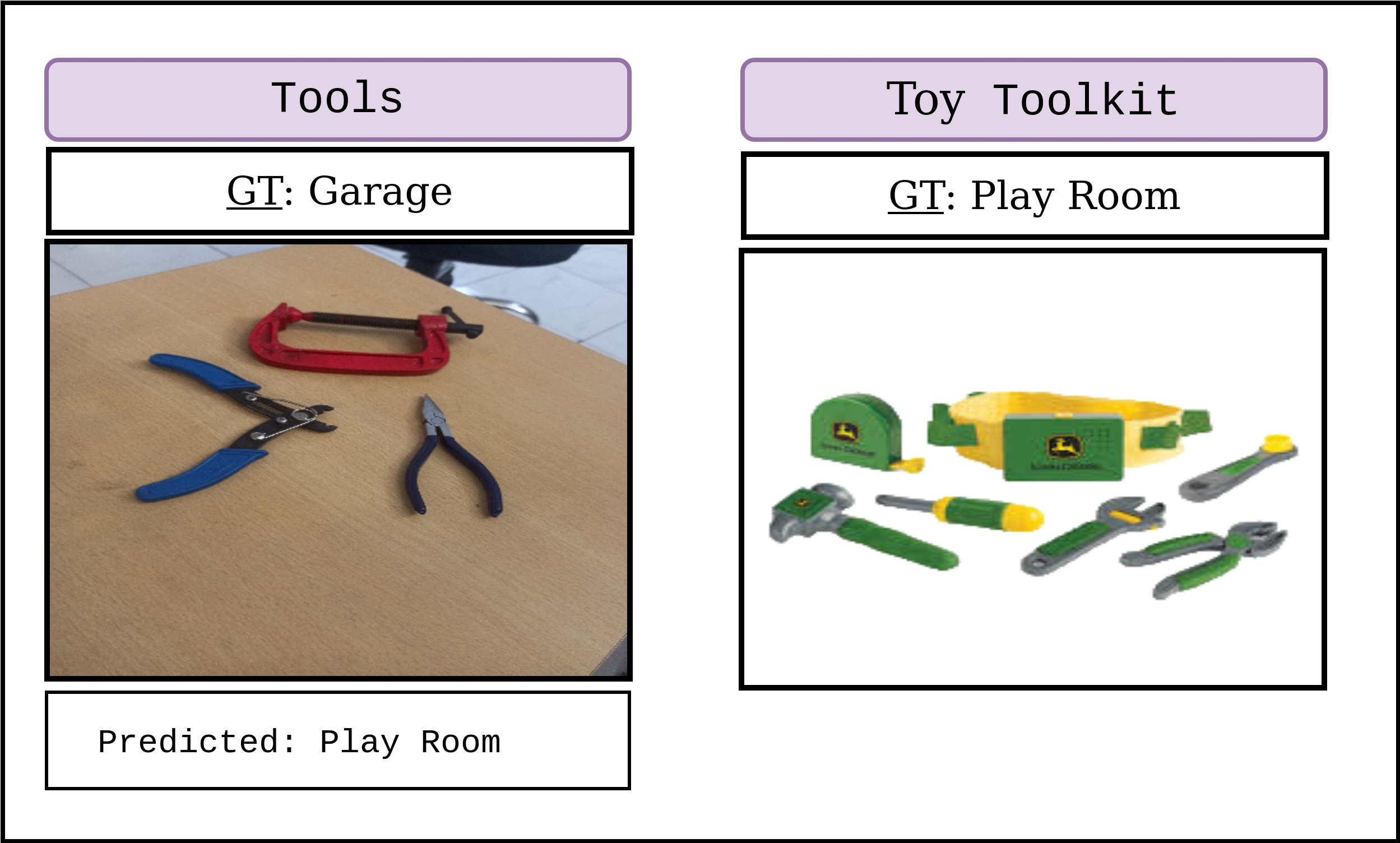}
    \caption{Failure with composite object categories; \textit{tools} was not a category in our training set, but they were incorrectly associated with the \textit{play room} because they were structurally similar to the \textit{toy toolkit} that was in the training set.}
    \label{figure:qual-failure:2-toy tools}
    \vspace{-1em}
\end{figure}

Figure~\ref{figure:qual-seen-success} shows the result of using our framework with images of previously seen objects but in noisy, previously unseen backgrounds. In each case, the object's room association was estimated correctly.
% images of extension of our results to real-world objects in four different scenarios, where the model is able to successfully classify the object to the correct room in each case.  
Next, Figure~\ref{figure:qual-success:handwash} shows the success cases when our trained framework was used with objects from previously unseen object categories. Success (i.e., estimating the correct room association for the objects) can be attributed to leveraging commonsense knowledge extracted from similar images.

% the  our model on unseen images, where the model is able to identify the correct room mapping solely on the basis of commonsense learnt from similar images. 
% Figures~\ref{figure:qual-fail:seen},

Figures~\ref{figure:qual-failure:1-air pods} and~\ref{figure:qual-failure:2-toy tools} show some examples of our framework's limitations. In Figure \ref{figure:qual-failure:1-air pods}, an input image of earpods (not present in the training set) was mapped to the \textit{utility room} because it was similar in appearance to hair dryers that were known to our framework. %  due to the presence of similar images of hair dryers and the model’s lack of understanding of scale.
Figure~\ref{figure:qual-failure:2-toy tools} shows another failure case in which our framework estimated the room association for actual tools (which it has not seen before) as \textit{playroom} because the training set contained an image of a toy tool kit in a playroom. However, each tool, when considered individually, is associated with the correct room location. These results support hypothesis \textbf{H4}. %when the training set contains images of a toy tool kit, even though the model correctly predicts the room when each tool is passed individually.

% \begin{figure}[tb]
%   \centering
%   \begin{minipage}[b]{\linewidth}
%     \includegraphics[width=\linewidth]{sections/figures/success-3(a).png}
%     \caption{Qualitative Results: Success Example for Unseen Object Category; Handwash,Plier[not in our train set] was predicted correctly}
%     \label{fig:untrained-tsne}
%   \end{minipage}\\
%   \vspace{0.57cm}
%   \begin{minipage}[b]{\linewidth}
%     \includegraphics[width=\linewidth]{sections/figures/fail-2.png}
%     \caption{Qualitative Results: Failure cases for Power Drill, Book [present in our train dataset]}
%     \label{fig:clustered-tsne}
%   \end{minipage}
% \end{figure}

%%%%%%%%%%%%%%%%%%%%%%%%%%%%%%%%%%%%%%%%%%%%%%%%%%%%%%%%%%%%%%%%%%%%%
%%%%%%%%%%%%%%%%%%%%%%%%%%%%%%%%%%%%%%%%%%%%%%%%%%%%%%%%%%%%%%%%%%%%%

% \input{sections/qualitative}
\section{Conclusion and Future Work}
Accurately estimating object-room affinities is an important step in performing scene rearrangement tasks. We presented a framework called CLIPGraphs, which estimates these affinities by leveraging the complementary strengths of commonsense knowledge, data-driven methods, and multimodal (vision, language) embeddings. Specifically, our framework encodes prior human preferences in a knowledge graph and considers CLIP-based image and language embeddings of nodes in this graph. It then uses Graph Convolutional Network (GCN)-based embeddings of these CLIP embeddings to learn and estimate the object-room affinities.
% We started with pre-existing CLIP vision-language embeddings and developed a novel graph convolutional network-based visual ranking module that generates affinities between various objects and rooms using multi-modal input embeddings and a knowledge graph generated from human preferences. Our 
We experimentally evaluated our framework's performance in estimating object-room affinities using our IRONA dataset of 8040 images of 268 benchmark object categories. We experimentally demonstrated a substantial improvement in the ability to estimate object-room affinities compared with language encoder embeddings and the GPT-3 LLM. We also showed qualitatively that our framework provides robustness to previously unseen noisy backgrounds. % a high increase in both metrics as compared to language encoder baselines. Further, Tables \ref{table:CLIP-untrained} and \ref{table:our-model} supports our claim that our graph based model learns to predict the object-room affinities much more accurately than existing off the shelf CLIP Embeddings. The fact that we achieve comparable results in Tables \ref{table:lang-claim}, \ref{table:our-model} when given CLIP language embeddings instead of CLIP Image embeddings during inference supports our multimodality claim and gives our model an additional edge over existing works. Our qualitative results highlighting our model's performance on real-world noisy images and unseen object categories imply that our model is generalizable. These results and the stellar $20\%$ gain over the best baseline prove that CLIPGraphs can be used to improve visual commonsense reasoning for robotic tasks involving navigation and rearrangement.

Our framework opens up directions for further research. For example, we plan to train our model with top-$k$ correct rooms to generate object-room affinities that would be useful in downstream tasks such as multi-object navigation.% where we humans use the semantic and utility-based organization of how we arrange our houses. 
We also plan to develop personalized or task-specific embeddings that allow our framework to calculate object-room affinities tailored to individual users, homes, or tasks. This will enable physical robots to assist humans in complex scene rearrangement tasks, and other embodied AI tasks characterized by semantic organization.% We will also investigate learning user-specific or task-specific embeddings, enabling the framework to compute object-room affinities that adapt to the particular person, home, or task under consideration. The long-term objective is to enable physical robots to assist humans in scene rearrangement tasks in complex domains. %Additionally, we plan to make this customizable to a particular user. This includes observing user behavior regarding object-room preferences and fine-tuning the embeddings to be useful for that particular user.    

%%%%%%%%%%%%%%%%%%%%%%%%%%%%%%%%%%%%%%%%%%%%%%%%%%%%%%%%%%%%%%%%%%%%%
%%%%%%%%%%%%%%%%%%%%%%%%%%%%%%%%%%%%%%%%%%%%%%%%%%%%%%%%%%%%%%%%%%%%%

\bibliography{references}
% \input{sections/ablations}
%%%%%%%%%%%%%%%%%%%%%%%%%%%%%%%%%%%%%%%%%%%%%%%%%%%%%%%%%%%%%%%%%%%%%
%%%%%%%%%%%%%%%%%%%%%%%%%%%%%%%%%%%%%%%%%%%%%%%%%%%%%%%%%%%%%%%%%%%%%

\end{document}